\documentclass{article}
\usepackage{spconf,amsmath,graphicx,hyperref}

\usepackage{amssymb}
\usepackage{adjustbox}
\usepackage{color}
\usepackage[dvipsnames]{xcolor}

\title{Adaptive and Balanced Re-initialization for Long-timescale Continual Test-time Domain Adaptation}
\name{\parbox{\linewidth}{\centering
Yanshuo Wang$^{1,2,3}$ \quad Jinguang Tong$^{4}$ \quad Jun Lan$^{3}$ \quad Weiqiang Wang$^{3}$ \quad Huijia Zhu$^{3}$ \\ Haoxing Chen$^{*,3}$ \quad Xuesong Li$^{*,5,4}$ \quad Jie Hong$^{*,6}$}\thanks{* Corresponding author}}
\address{$^{1}$Hong Kong Polytechnic University $^{2}$Eastern Institute for Advanced Study $^{3}$Ant Group \\ $^{4}$Australian National University $^{5}$CSIRO $^{6}$The University of Hong Kong}

\begin{document}
%
\maketitle
\begin{abstract}
Continual test-time domain adaptation (CTTA) aims to adjust models so that they can perform well over time across non-stationary environments. While previous methods have made considerable efforts to optimize the adaptation process, a crucial question remains: Can the model adapt to continually changing environments over a long time? In this work, we explore facilitating better CTTA in the long run using a re-initialization (or reset) based method. First, we observe that the long-term performance is associated with the trajectory pattern in label flip. Based on this observed correlation, we propose a simple yet effective policy, Adaptive-and-Balanced Re-initialization (ABR), towards preserving the model's long-term performance. In particular, ABR performs weight re-initialization using adaptive intervals. The adaptive interval is determined based on the change in label flip. The proposed method is validated on extensive CTTA benchmarks, achieving superior performance.
\end{abstract}
\begin{keywords}
Domain adaptation, continual test-time domain adaptation, long-timescale continual test-time domain adaptation
\end{keywords}

\vspace{-12pt}
\section{Introduction}
\label{sec:intro}
Humans and animals have remarkable abilities to learn and adapt to new environments or conditions.
For instance, dogs can master complex commands and execute tasks with precision. They also demonstrate adaptability by adjusting to changing environments while maintaining their ability to follow commands.
Now, researchers seek to empower machine learning algorithms to adapt to changing domains.

\begin{figure}[t]
\centering
\includegraphics[width=1.0\linewidth]{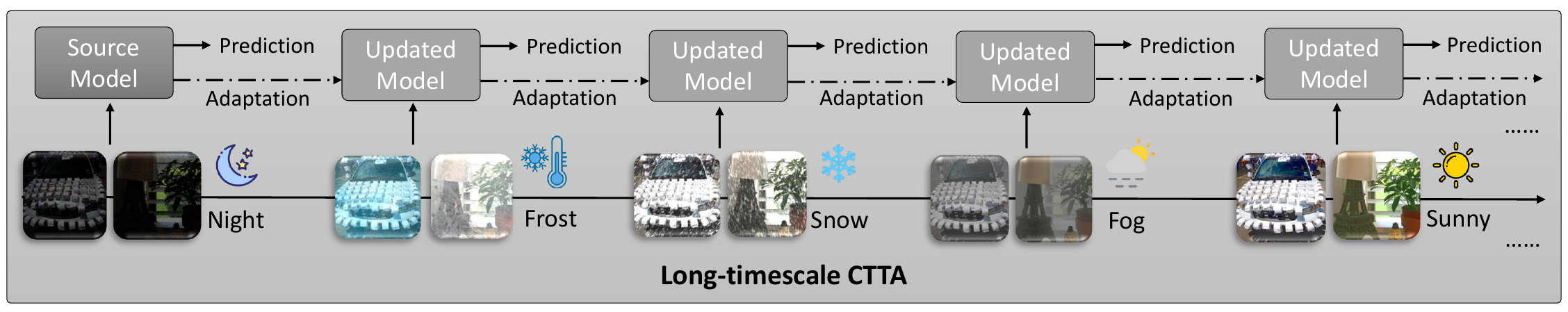}
\vspace{-24pt}
\caption{Long-timescale CTTA. The source model is adapted to a large number of continually changing domains. The timestep corresponds to the number of test-time samples.}
\vspace{-10pt}
\label{fig:ctta}
\end{figure}

A number of previous studies focus on domain adaptation (DA), which adapts a model pre-trained on the source domain to different domains.
The methods in DA are categorized into two types: the alignment of the feature distribution through discrepancy losses between domains~\cite{long2015learning} 
and adversarial training~\cite{tzeng2017adversarial}.
The focus has recently shifted to a more realistic setting known as continual test-time domain adaptation (CTTA), which involves adapting models in non-stationary conditions at test time~\cite{wang2022continual, wang2024backpropagation, wang2025dynamic}. 
Although existing CTTA methods have made considerable improvements in optimizing the model during the adaptation process~\cite{niu2022efficient, wang2024continual}, 
most of them overlook the long-term adaptation capability of the models. Recently, in~\cite{press2024rdumb}, a set of CTTA methods has been evaluated across large benchmarks. Most methods were found to fail in maintaining consistent performance over a prolonged period. In some extreme cases, their performances even collapse. Similar to~\cite{press2024rdumb}, in this study, we focus on addressing long-timescale continual test-time domain adaptation (long-timescale CTTA), as shown in Fig.~\ref{fig:ctta}.

Through toy experiments, we find that, to some extent, the model's long-term performance correlates with the label flip~\cite{toneva2018empirical, press2024entropy}, 
which measures the difference between the current and previous-time models.
It is observed that performance loss decreases when we re-initialize (or reset) the model near the time point when the label flip rises rapidly. The label flip pattern appears to provide a useful signal for when to re-initialize model weights.

Based on this observation, we propose an adaptive weight re-initialization policy for long-term performance preservation, based on the label-flip trajectory, named Adaptive-and-Balanced Re-initialization (ABR). 
Specifically, ABR adaptively selects the timing of the model re-initialization based on the label-flipping trajectory. The re-initialization comprises restoring the source model weights and shrinking the current model weights using a suitable ratio, aiming to achieve balanced effects from both the source and the current model. The contributions of this work can be summarized as follows: 
1) We find that changes in label flip could be an explicit signal to model re-initialization for reducing performance loss in the long-term model adaptation process.
2) We propose a straightforward policy, ABR, which provides a balanced re-initialization of model weights at adaptive timings, effectively preserving the model performance. The timing of the re-initialization depends on the label flip.

\section{Methodology}
\label{sec:method}
In this section, we discuss the model's long-term performance on long-timescale CTTA. We find that the label flip is strongly associated with long-term performance. Then we propose a solution, ABR, to preserve the long-term performance.

\subsection{Long-term Performance}
\begin{figure}[t]
\centering
\includegraphics[width=1.0\linewidth]{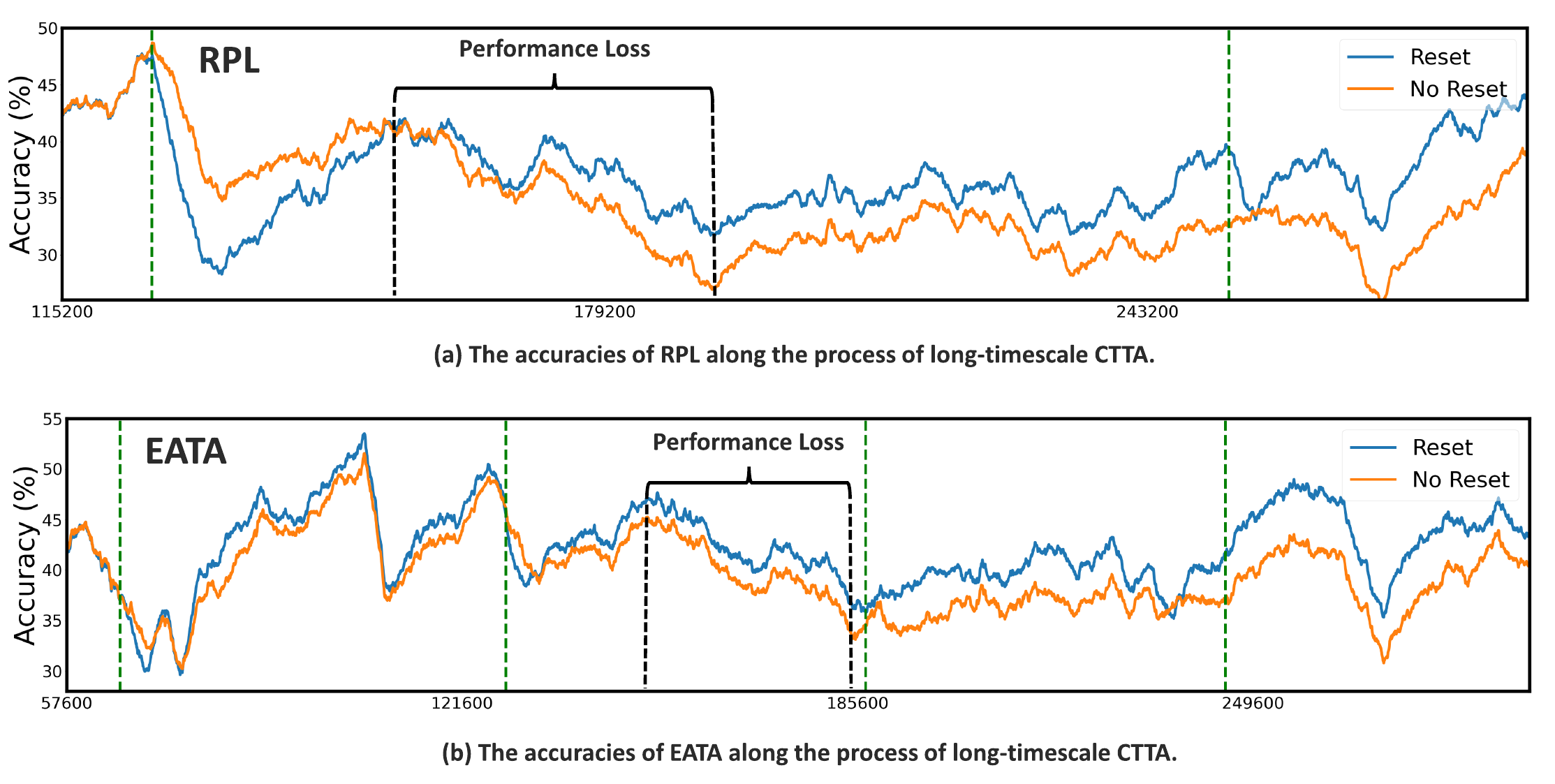}
\vspace{-24pt}
\caption{Performance in long-timescale CTTA. As shown in the sub-figure below, test data continually changes from one domain to another. The classification accuracies of RPL~\cite{rusak2021selflearning} and EATA~\cite{niu2022efficient} on the CCC dataset~\cite{press2024rdumb} are plotted in the upper and middle sub-figures, respectively. As shown in the figure, methods with a reset policy, represented by the \textcolor{NavyBlue}{blue} line, achieve superior performance compared to those without a reset policy, represented by the \textcolor{orange}{orange} line. Moreover, this advantage becomes more distinct in the later run. The dotted \textcolor{LimeGreen}{green} line represents the reset time point.}
\vspace{-12pt}
\label{fig:plasicity_loss1}
\end{figure}

The reason behind reduced long-term performance is error accumulation resulting from either entropy minimization~\cite{wang2021tent} or pseudo-labeling~\cite{goyal2022test, wang2022continual}. Those methods amplify errors and overfit during adaptation, reducing the model's ability to learn new domain samples. Unfortunately, most CTTA methods are developed using those techniques and heavily rely on them to adapt the model to a new domain. As shown in Fig.~\ref{fig:plasicity_loss1}, two tested CTTA models, RPL~\cite{rusak2021selflearning} and EATA~\cite{niu2022efficient}, based on entropy minimization, clearly degrade in performance. 
Even EATA, which uses regularization, still struggles to maintain classification accuracy in the later runs.

\begin{figure}[t]
\centering
\includegraphics[width=0.99\linewidth]{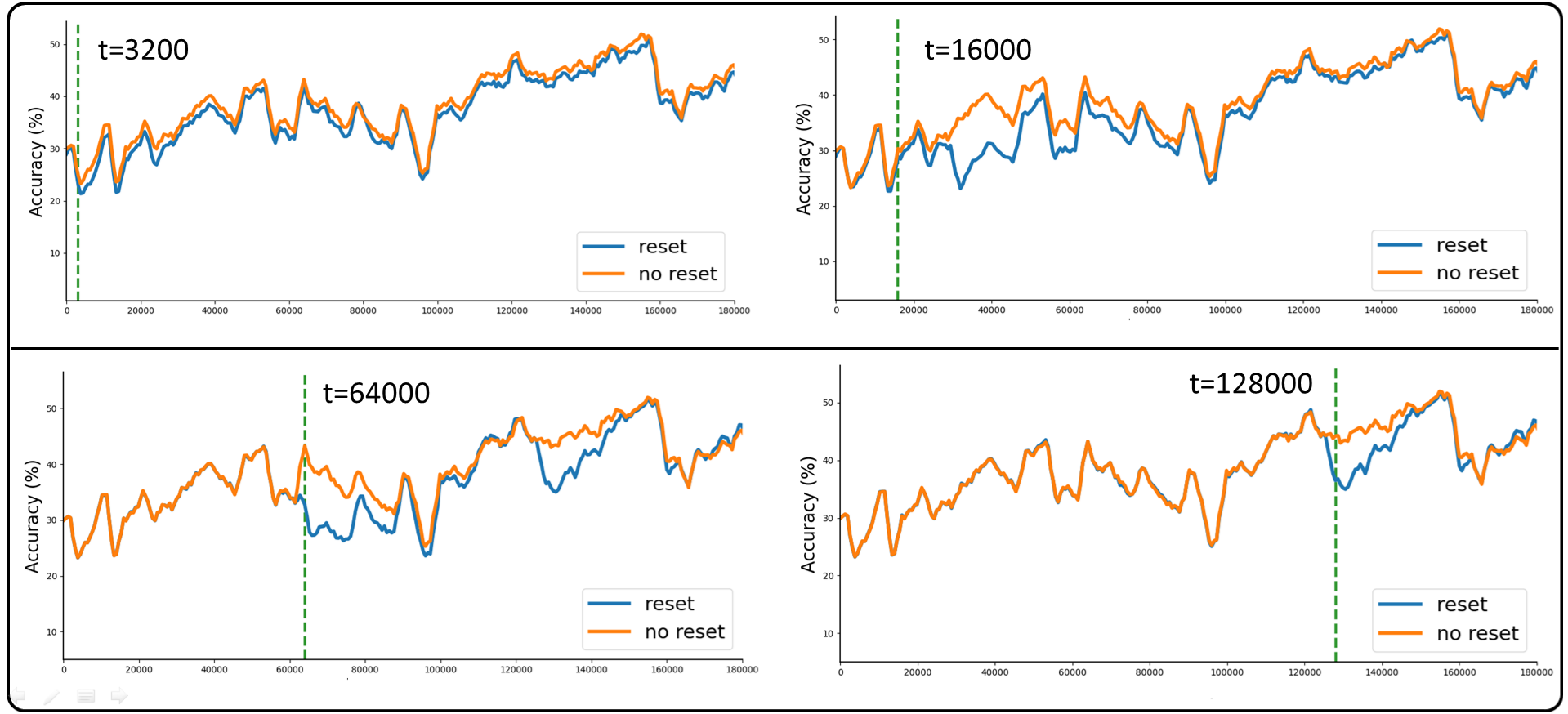}
\vspace{-10pt}
\caption{Model long-term performance under different reset timings. Using RPL~\cite{rusak2021selflearning}, we trigger the reset using four randomly chosen timings (the \textcolor{NavyBlue}{blue} line). 
We also use the no-reset policy for comparison (the \textcolor{orange}{orange} line). The dotted \textcolor{LimeGreen}{green} line represents the reset time point. We can see from the figure that these four reset timings do not yield better performance, indicating that suitable re-initialization times are needed.}
\vspace{-12pt}
\label{fig:reset_interval}
\end{figure}

\noindent \textbf{When to Trigger the Re-initialization?}
A natural question arises: when should model weight re-initialization be triggered to preserve performance during long-timescale CTTA?
To explore this, we conduct another simple experiment in which RPL~\cite{rusak2021selflearning} is evaluated on long-timescale CTTA. The reset timing is randomly set at $3.2$K, $16$K, $64$K, and $128$K. The results, shown in Fig.~\ref{fig:reset_interval}, indicate that randomly triggering the re-initialization does not yield superior performance over a long time. 
One of the existing methods~\cite{press2024rdumb}, RDumb, resets the model by a fixed time interval tuned using the related validation dataset. 
However, using external data limits the realistic usage of the model.
Moreover, extending such re-initialization with a tuned interval to other CTTA methods can be difficult, since different methods have different optimal values.
Thus, we propose an adaptive approach to determine the optimal time.

\noindent \textbf{Label Flip.}
\label{subsec:LF}
During model adaptation, a prediction difference would occur in the current-time domain or environment between the current adapted model and the model at the previous time. 
More specifically, we predict a set of test images using the models at the current and last time of adaptation. After passing these images to both models, we can record the predicted classes of each model. We refer to the predicted class differences between the two models as the label flip~\cite{toneva2018empirical, press2024entropy} 
at the current time.

\begin{figure}[t]
\centering
\includegraphics[width=0.99\linewidth]{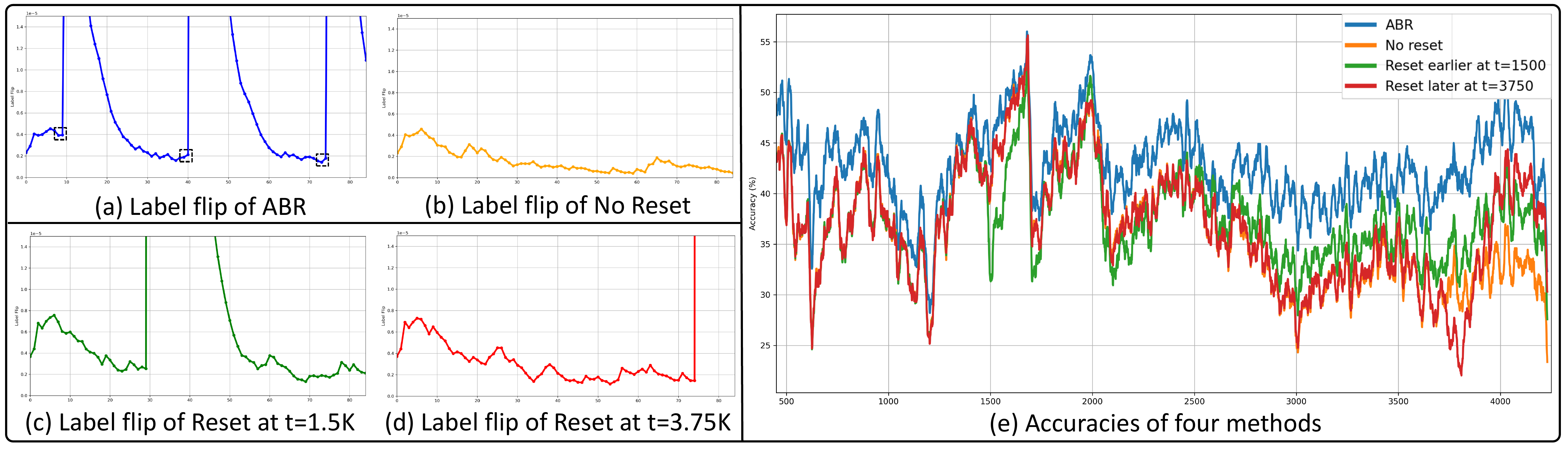}
\vspace{-10pt}
\caption{Label flip and long-term performance in long-timescale CTTA. The black dot box highlights the timing of re-initialization.}
\vspace{-10pt}
\label{fig:label_flip}
\end{figure}

In this study, we observe an interesting phenomenon regarding the relationship between label flip and long-term performance loss. If we perform re-initialization immediately after the sharp increase in the label flip, performance would be better preserved during later adaptation. In contrast, triggering re-initialization at other time points would not lead to improvements in maintaining long-term performance. 
We carry out experiments to illustrate this point, which is shown in Fig.~\ref{fig:label_flip}. We use the RPL as the baseline, with four re-initialization times: adaptive timing of the proposed ABR, no re-initialization, re-initialization at an early time, and re-initialization at a late time. As shown in the figure, initiating the reset immediately after the rise in label flip yields significantly better performance than starting later.

\subsection{Adaptive-and-Balanced Re-initialization}
\begin{figure}[t]
\centering
\includegraphics[width=0.99\linewidth]{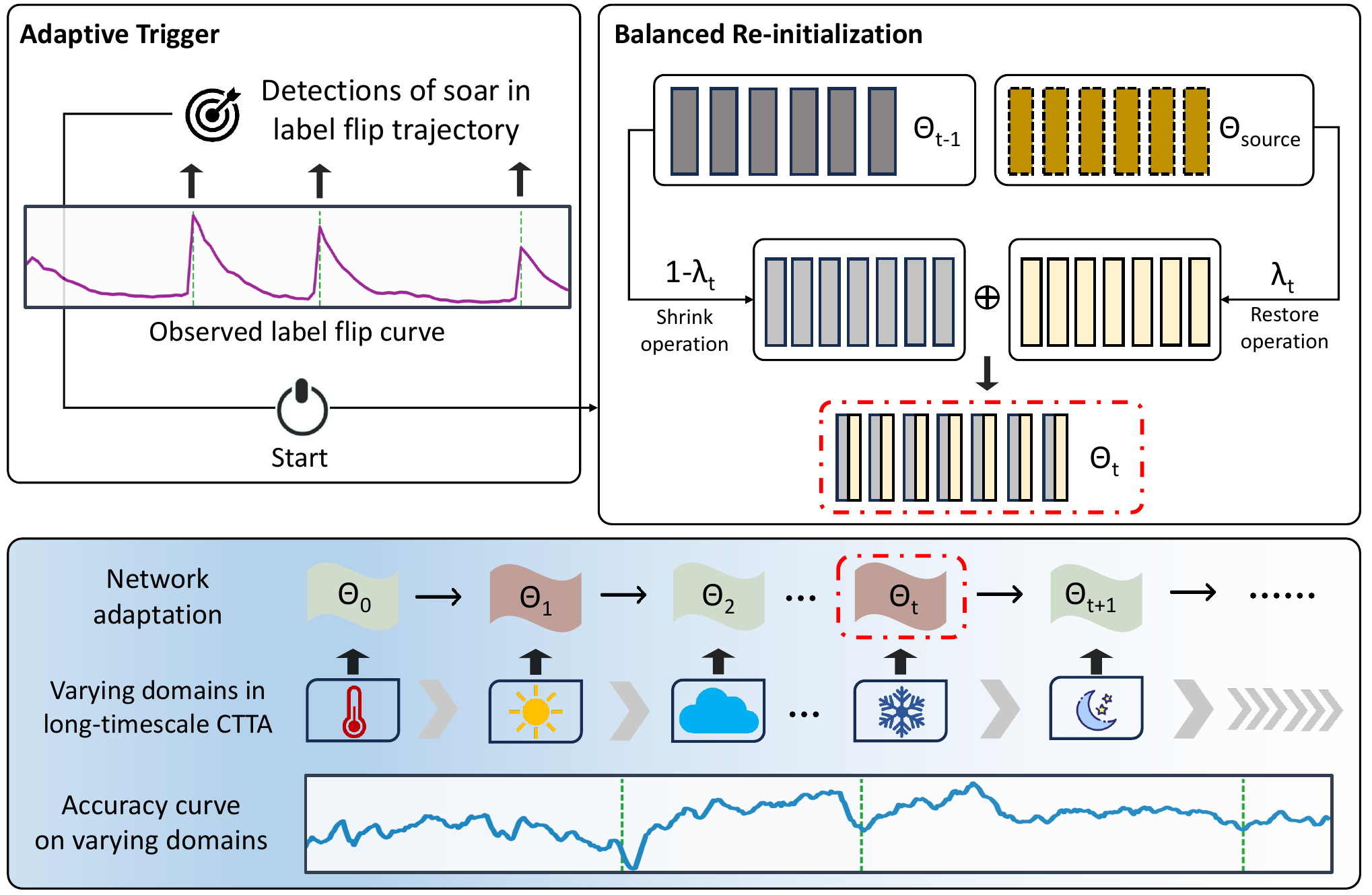}
\vspace{-10pt}
\caption{The framework of the proposed Adaptive-and-Balanced Re-initialization (ABR). When the model continuously adapts in non-stationary domains, balanced re-initialization is triggered adaptively to mitigate long-term performance loss. For the balanced re-initialization, we use the strategy called "shrink-restore" to keep parts of previously learned knowledge of the adapted model while restoring knowledge from the source model.}
\vspace{-10pt}
\label{fig:framework}
\end{figure}

To mitigate the model's performance loss during long-timescale CTTA, we propose an adaptive re-initialization policy, ABR, whose overall framework is shown in Fig.~\ref{fig:framework}.

\noindent \textbf{Label Flip as Signal.}
To better quantify label flip, we use a metric similar to that of \cite{press2024entropy}, based on the model's confidence at the previous time and the increased confidence at the current time for the changed predicted class. 
Here, we have the raw label flip at time $t$:
\begin{equation}\label{eq:label_flip}
\text{LF}_{raw,t} = \sum_{i} \mathbb{I}(i) \cdot c_{i,t} \cdot (c_{i,{t}} - c_{i,{t-1}})
\end{equation}
where $\mathbb{I}(.)$ is an indicator function that denotes whether the label flip (\textit{i.e.}, changes in predicted class) exists. $c_{i, t}$ represents the predicted confidence of the current-time model, while $c_{i,t-1}$ is the one from the model at the previous time. 
Along with adaptation, varying environments cause severe fluctuations in the trajectory of the flip label, making it difficult to use.
Thus, we process the raw data of label flip via an exponential moving average to get a smooth trajectory:
\begin{equation}\label{eq:mv_average}
\text{LF}_{t}=\alpha \text{LF}_{t-1} + (1-\alpha)\text{LF}_{raw,t}
\end{equation}
where $t$ is the time and $\alpha$ is the ratio for updating the \text{LF} along the adaptation. 
Throughout the experiments, we set $\alpha$ to $0.5$.
The operation in Eq.~(\ref{eq:mv_average}) ensures that label flip can provide a stable signal to the adaptive trigger.

\noindent \textbf{Adaptive Trigger.}
Since label flip relates to maintaining the long-term performance during the long-timescale CTTA, as discussed in subsection~\ref{subsec:LF}, we choose to trigger the re-initialization at the time based on the changes in label flip trajectory.
Specifically, we aim to identify the time point at which the label flip rate increases rapidly. 
To detect this point, we compute the slope between the minimum point of label flip and its value at the current time. The reason is that, at the minimum point, the slope of the label flip curve shows an obvious increase as the curve shifts from stable to rising. A positive, large slope indicates that the curve is moving upward quickly.

We define the minimum point, $\text{LF}_{\min}$, of the label flip trajectory by finding the lowest value along the trajectory:
\begin{equation}
\text{LF}_{\min} = \min (\text{LF})
\end{equation}
where $\text{LF}$ is the set that includes all values of label flip up to time $t$. The minimum point $\text{LF}_{\min}$ is regarded as a reference point. The difference between $\text{LF}_t$ and this reference point should be measured to see if or not there is a rising trend. Therefore, we can compute the slope between two points at time $t$, $\text{S}_t$:
\begin{equation}\label{eq:slope}
\text{S}_t = \frac{\Delta \text{LF}}{\Delta t}
\end{equation}
where $\Delta \text{LF}=\text{LF}_t-\text{LF}_{\min}$, $\Delta t=t-t_{\min}$, and $t_{\min}$ is the time of $\text{LF}_{\min}$.

Finally, we introduce $\beta$ as a slope threshold to help identify the clear rise in the label-flip trajectory. We give the condition of the trigger of re-initialization as follows:
\begin{equation}\label{eq:detection}
\text{S}_t > \frac{\beta}{\sqrt{t - t_{\min}}} 
\end{equation}\label{equ:alpha_threshold}
where the threshold $\beta$ is set to $2e-6$ for all experiments. Note that this trajectory sometimes remains stable for a long time.
When the duration of the stable period is large, the slope tends to be smaller.
Hence, we dynamically scale the threshold $\beta$ using the square root of the duration of the stable period. We aim to lower the threshold accordingly to avoid missing obvious upward trends.
The inequation in Eq.~(\ref{eq:detection}) determines the time of re-initialization. After the trigger is activated, the balanced re-initialization will be executed.

\noindent \textbf{Balanced Re-initialization.}
Followed by the adaptive trigger, the simplest way of weight re-initialization is to reset the model directly (see Fig.~\ref{fig:plasicity_loss1}).
This involves assigning the model a weight entirely derived from the pre-trained model of the source domain~\cite{press2024rdumb}.
However, this would cause the model to lose all the previous knowledge it had acquired. 
We suppose retaining some of the model's past knowledge at each weight re-initialization time would benefit long-term performance preservation.

To dynamically balance learned and restored source knowledge, we use the label flip to guide an adaptive re-initialization process. If the label flip value triggered by the adaptive trigger is significantly larger than the minimum point, it indicates a considerable deviation. That means that the model requires more of the source model weights to maintain long-term performance. Conversely, if the value is smaller, more previously learned knowledge can be retained.
Therefore, we design the weight update rules as follows: 
\begin{equation}\label{eq:adptive_plasitics}
\theta_t = \lambda_t \theta_\text{source} + (1- \lambda_t) \theta_{t-1}
\end{equation}
where $\theta_{\text{source}}$ is the source model weight and $\theta_t$ is the model weight at time $t-1$. $\lambda_t$ is the dynamic adjusting coefficient, and it is expressed as: 
\begin{equation}
\lambda_t = \frac{\text{LF}_t}{\text{LF}_t + \text{LF}_{\min}}
\end{equation}
where $\lambda_t$ adaptively determines the degree to which the source model weight is restored or the adapted model weight is shrunk. The re-initialization method in Eq.~(\ref{eq:adptive_plasitics}) better ensures that the adapted model learns comprehensive knowledge in a balanced way.

\section{Experiments}
\label{sec:expt}
To test the effectiveness of the proposed ABR, extensive long-timescale CTTA experiments are conducted on three large and comprehensive datasets: CIN-C~\cite{hendrycks2019benchmarking}, CIN-3DCC~\cite{kar20223d}, and CCC~\cite{press2024rdumb}. The ablation studies are also provided.

\subsection{Implementation}
Following RDumb~\cite{press2024rdumb}, we use the pre-trained ResNet-50~\cite{he2016deep} as the default adaptation model. In addition, our approach is built on the EATA framework~\cite{niu2022efficient}, as RDumb does. In all experiments, we use a batch size of $64$.
Note that the results for CIN-C and CIN-3DCC are obtained by averaging over $10$ different permutations of corruptions, and the results for CCC are also obtained by averaging across its combinations with different seeds and transition speeds. For CIN-C and CIN-3DCC, we follow the settings in~\cite{press2024rdumb}, which use the highest severity level $5$ as the default. The threshold $\beta$ is set to $2e-6$ for across all datasets.

\subsection{Main Results}
To demonstrate the advantages of ABR in long-timescale CTTA, we compare it with several existing methods across $3$ datasets: BatchNorm (BN)~\cite{schneider2020improving}, Test Entropy Minimization (TENT)~\cite{wang2021tent}, Robust Pseudo-Labeling (RPL)~\cite{rusak2021selflearning}, Soft Likelihood Ratio (SLR)~\cite{mummadi2021test}, Conjugate Pseudo Labels (CPL)~\cite{goyal2022test}, Continual Test Time Adaptation (CoTTA)~\cite{wang2022continual}, Efficient Test Time Adaptation (EATA)~\cite{niu2022efficient}, EATA without Weight Regularization (ETA)~\cite{niu2022efficient}, and RDumb \cite{press2024rdumb}.

The main results are provided in Tab.~\ref{tab:main_tab}. The table shows that most CTTA baselines struggle to maintain consistent performance over the long term during adaptation. In some cases, their performance deteriorates so much that it falls below the performance of the pre-trained model, indicating a clear failure to adapt effectively to new information. The SLR's accuracy is notably worse than the pre-trained model's on all datasets. The situation becomes even more severe in the most challenging dataset, CCC, where many methods completely collapse. For instance, CPL achieves only $0.14\%$ accuracy on CCC-Hard. This suggests that these methods do not preserve performance well enough to effectively adapt to the data from the new domain, highlighting a limitation in their ability to handle continual adaptation.

Overall, among all baseline methods, only EATA and RDumb exhibit some degree of robustness against long-term performance loss, either through simple resetting or regularization. In particular, RDumb~\cite{press2024rdumb} performs relatively well, with the reset time interval tuned from a separate validation set. 
However, tuning the time interval is time-consuming work. Moreover, an inappropriate interval may lead to more performance loss than the no-reset method, as shown in Fig.~\ref{fig:reset_interval}.
Our proposed method, ABR, which involves an adaptive trigger of balanced re-initialization operation, surpasses other adaptation methods across all datasets.
For example, ABR achieves a mean accuracy of $40.2\%$, improving by $2.3\%$ over the best baseline, RDumb. More importantly, the improvement becomes clearer when we use more difficult datasets. This further indicates that the proposed method can better handle varying domains in long-term processes.

\begin{table}[t]
\centering
\renewcommand{\arraystretch}{1.3}
\begin{adjustbox}{width=0.48\textwidth} 
    \begin{tabular}{l|ccccc|c}
    \hline
    \textbf{Method} & \textbf{CIN-C} & \textbf{CIN-3DCC} & \textbf{CCC-Easy} & \textbf{CCC-Medium} & \textbf{CCC-Hard} & \textbf{Average} \\ \hline
    \textbf{Pre-trained} & 18.0 ± 0.0 & 31.5 ± 0.22 & 34.1 ± 0.22 & 17.3 ± 0.21 & 1.5 ± 0.02 & 20.5 \\
    \textbf{BN}~\cite{schneider2020improving} & 31.5 ± 0.02 & 35.7 ± 0.02 & 42.6 ± 0.39 & 27.9 ± 0.74 & 6.8 ± 0.31 & 28.9 \\ 
    \textbf{TENT}~\cite{wang2021tent} & 15.6 ± 3.5 & 24.4 ± 3.5 & 3.9 ± 0.58 & 1.4 ± 0.17 & 0.51 ± 0.07 & 9.2 \\
    
    \textbf{RPL}~\cite{rusak2021selflearning} & 21.8 ± 3.6 & 30.0 ± 3.6 & 7.5 ± 0.83 & 2.7 ± 0.36 & 0.67 ± 0.14 & 12.5 \\ 
    \textbf{SLR}~\cite{mummadi2021test} & 12.4 ± 7.7 & 12.2 ± 7.7 & 22.2 ± 18.4 & 7.7 ± 9.0 & 0.66 ± 0.57 & 11.0 \\ 
    \textbf{CPL}~\cite{goyal2022test} & 3.0 ± 3.3 & 5.7 ± 3.3 & 0.41 ± 0.06 & 0.22 ± 0.03 & 0.14 ± 0.01 & 1.9 \\
    \textbf{CoTTA}~\cite{wang2022continual} & 34.0 ± 0.68 & 37.6 ± 0.68 & 14.9 ± 0.88 & 7.7 ± 0.43 & 1.1 ± 0.16 & 19.1 \\
    
    \textbf{EATA}~\cite{niu2022efficient} & 41.8 ± 0.98 & 43.6 ± 0.98 & 48.2 ± 0.6 & 35.4 ± 1.0 & 8.7 ± 0.8 & 35.5 \\ 
    \textbf{ETA}~\cite{niu2022efficient} & 43.8 ± 0.33 & 42.7 ± 0.33 & 41.4 ± 0.95 & 1.1 ± 0.43 & 0.23 ± 0.05 & 25.8 \\ 
    \textbf{RDumb}~\cite{press2024rdumb} & 46.5 ± 0.15 & 45.2 ± 0.15 & 49.3 ± 0.88 & 38.9 ± 1.4 & 9.6 ± 1.6 & 37.9 \\ 
    \textbf{TCA}~\cite{ni2025maintaining} & - & - & 49.1 ± 0.35 & 39.5 ± 0.53 & 10.1 ± 0.22 & - \\ \hline
    
    \textbf{ABR (ours)} 
     & \textbf{47.5 ± 0.26} & \textbf{45.8 ± 0.10} & \textbf{51.4 ± 0.83} & \textbf{43.7 ± 1.21} & \textbf{12.7 ± 0.72} & \textbf{40.2} \\  
    \hline
    \end{tabular}
\end{adjustbox}
\caption{Performances of different CTTA methods across three benchmarks, CIN-C, CIN-3DCC, and CCC, with multiple difficulties. We report the mean accuracy in $\%$ with its standard deviation.}\label{tab:main_tab}
\end{table}

\section{Conclusion}
\label{sec:conclusion}
In this work, we find that label flipping monitoring provides a reliable signal of performance loss during long-timescale adaptation. Based on the observation, we propose ABR, a straightforward yet effective policy for preserving the long-term performance through adaptive re-initialization of model weights. Our experiments show that ABR significantly enhances the model's ability to maintain classification accuracy in long-timescale CTTA. Moreover, ABR can be integrated into most CTTA methods to maintain their continual long-term adaptation performances.

\bibliographystyle{IEEEbib}
\bibliography{strings,refs}

\end{document}